\documentclass{article}

 \usepackage[preprint]{neurips_2025}

\usepackage[utf8]{inputenc} 
\usepackage[T1]{fontenc}    
\usepackage[colorlinks=true, allcolors=blue]{hyperref}       
\usepackage{url}            
\usepackage{booktabs}       
\usepackage{amsfonts}       
\usepackage{amsmath}
\usepackage{xfrac}
\usepackage{nicefrac}       
\usepackage{microtype}      
\usepackage{xcolor}         
\usepackage{multirow}
\usepackage{lipsum}
\usepackage{subcaption}
\usepackage{graphicx}
\usepackage[capitalise,noabbrev]{cleveref}
\usepackage{threeparttable}
\usepackage{wrapfig}

\usepackage{circledsteps}
\pgfkeys{/csteps/inner color=white}
\pgfkeys{/csteps/fill color=orange}
\pgfkeys{/csteps/outer color=orange}

\title{Adaptive Block-Scaled Data Types}

\author{%
  Jack Cook\thanks{Massachusetts Institute of Technology\quad$^2$NVIDIA}
  \And
  Hyemin S. Lee$^1$
  \And
  Kathryn Le$^1$
  \And
  Junxian Guo$^1$
  \AND
  Giovanni Traverso$^1$
  \And
  \textbf{Anantha P. Chandrakasan}$^1$
  \And
  \textbf{Song Han}$^{1,2}$
  \AND
  \url{https://github.com/mit-han-lab/fouroversix}
}

\begin{document}

\maketitle

\begin{abstract}
  NVFP4 has grown increasingly popular as a 4-bit format for quantizing large language models due to its hardware support and its ability to retain useful information with relatively few bits per parameter.
  However, the format is not without limitations: recent work has shown that NVFP4 suffers from its error distribution, resulting in large amounts of quantization error on near-maximal values in each group of 16 values.
  In this work, we leverage this insight to design new \textit{Adaptive Block-Scaled Data Types} that can adapt to the distribution of their input values.
  For four-bit quantization, our proposed IF4 (Int/Float 4) data type selects between FP4 and INT4 representations for each group of 16 values, which are then scaled by an E4M3 scale factor as is done with NVFP4.
  The selected data type is denoted using the scale factor's sign bit, which is currently unused in NVFP4, and we apply the same insight to design formats for other bit-widths, including IF3 and IF6.
  When used to quantize language models, we find that IF4 outperforms existing 4-bit block-scaled formats, achieving lower loss during quantized training and achieving higher accuracy on many tasks in post-training quantization.
  We additionally design and evaluate an IF4 Multiply-Accumulate (MAC) unit to demonstrate that IF4 can be implemented efficiently in next-generation hardware accelerators.
\end{abstract}

\section{Introduction}

Reductions in numerical precision beget increases in matrix multiplication speed on modern hardware accelerators~\cite{wang_bfloat16_2019,micikevicius_fp8_2022,nvidia_pretraining_2026}.
As a result, many researchers have recently explored training large language models using 4-bit floating point numbers (i.e. FP4), which can be multiplied twice as fast as 8-bit numbers (i.e. FP8), or four times as fast as 16-bit numbers (i.e. FP16, BF16), on newer GPUs such as NVIDIA's B200 and AMD's MI355X~\cite{nvidia_pretraining_2026}.
However, to benefit from fast hardware support, both operands of a low-precision matrix multiplication must be quantized, meaning that during training, activations, weights, and gradients would all need to be quantized.
This making training with 4-bit numbers particularly challenging, as a 4-bit format can only represent 16 values.
For FP4, these are $\pm\{0, 0.5, 1, 1.5, 2, 3, 4, 6\}$.

\begin{figure}
    \centering
    \includegraphics[width=\linewidth]{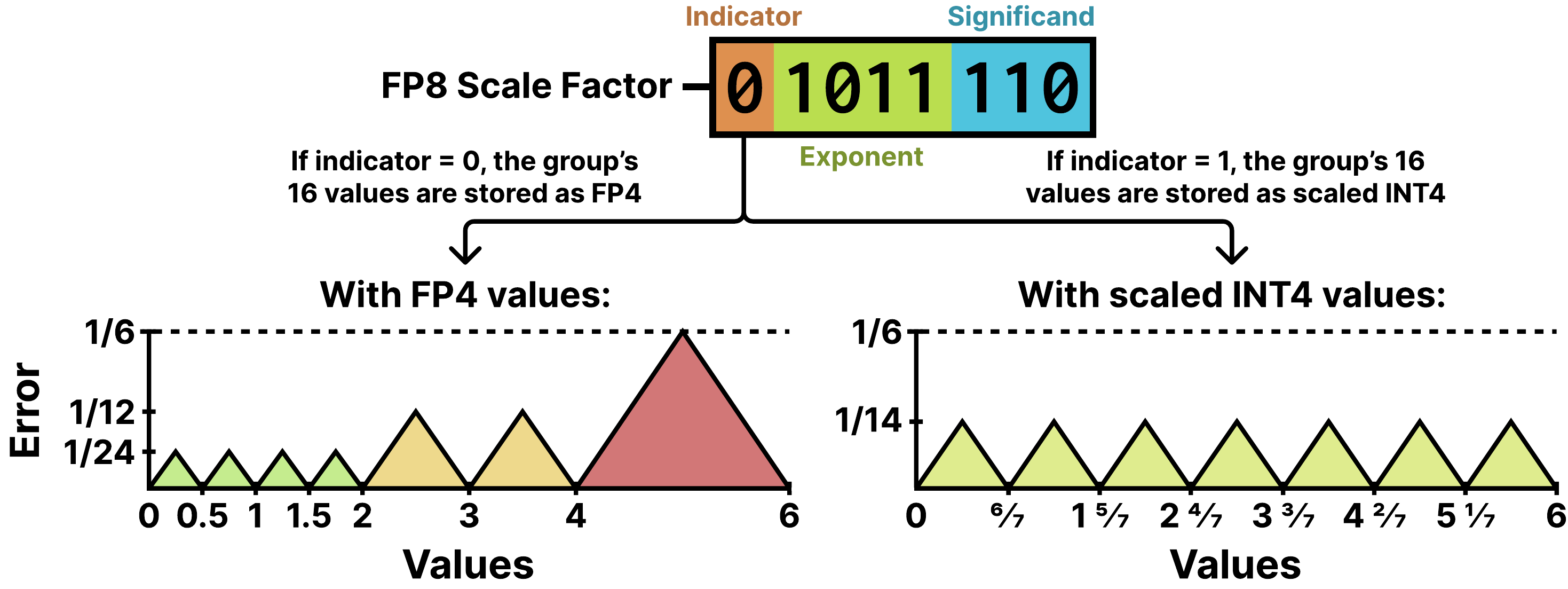}
    \caption{A visual explanation of the IF4 data type. Like NVFP4, individual values are scaled by an FP8 scale factor, each of which is shared by 16 values. However, since each value has its own sign bit, in NVFP4 all scale factors are positive and the scale factor's sign bit is never used. We repurpose this bit into an indicator which reflects whether the group's values are stored as FP4 or scaled INT4.}
    \label{fig:if4}
\end{figure}

Despite these challenges, many researchers have started training large language models with FP4~\cite{tseng_training_2025,castro_quartet_2026,chmiel_fp4_2025,nvidia_pretraining_2026}.
To work around the format's lack of precision, several operations have become essential parts of successful FP4 training recipes.
First, most works use block-scaled formats such as NVFP4 or MXFP4, which scale groups of FP4 values (containing 16 or 32 values respectively) by separately-quantized FP8 scale factors, limiting each individual group to the range of FP4 but allowing a significantly larger range of values across an entire tensor~\cite{rouhani_ocp_2023,nvidia_pretraining_2026}.
Other key operations include the random Hadamard transform which smooths outliers~\cite{tseng_training_2025}, stochastic rounding which provides unbiased estimates of gradients, and keeping many layers and operations in high precision which are too sensitive to quantize~\cite{nvidia_pretraining_2026}.
Each operation adds overhead, cutting into the performance gains that were offered by low-precision matrix multiplication in the first place.
As a result, each operation needs to be extremely efficient.
For 4-bit training to be viable, the combined overhead from all of these operations must be less than the time saved by using FP4 rather than FP8, which is roughly equivalent to the time it takes to do the FP4 matrix multiplication.

\begin{wrapfigure}{r}{0.5\textwidth}
    \includegraphics[width=\linewidth]{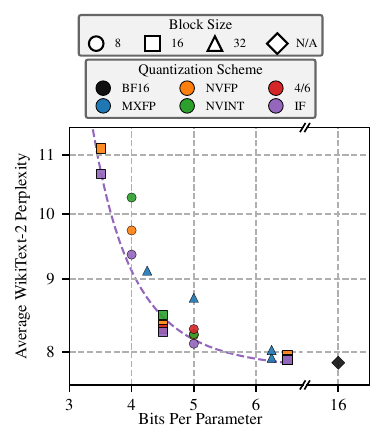}
    \caption{Adaptive Block-Scaled Data Types (IF) provide better model performance for most low-precision memory constraints (\cref{sec:alternative_data_types}).}
    \label{fig:ppl_vs_bits}
\end{wrapfigure}

Another recently proposed operation for NVFP4 training is Four Over Six (4/6), which reduces NVFP4 quantization error by adaptively scaling each group of FP4 values to a maximum value of either 4 or 6, rather than scaling all groups to the default maximum value of 6.
If a group is scaled to a maximum value of 4, the distribution of quantization error across the group changes in a way that favors near-maximal values in each group~\cite{cook_four_2026}.
However, while selecting between these scales improves downstream performance, using a maximum of 4 requires giving up two FP4 values, 6 and -6, when there are only 16 values to begin with.
Furthermore, since groups are scaled to 4 by multiplying the group's scale factor by 1.5, the NVFP4 global tensor scale needs to be reduced in order to prevent overflow, reducing the quantized tensor's dynamic range.
While it is clear that adaptive scaling is beneficial in block-scaled numerical formats, it is likely that these issues could be avoided with changes to the underlying format.

In this work, we introduce \textit{Adaptive Block-Scaled Data Types} to enable more accurate low-precision training and inference.
In NVFP4 with Four Over Six, scaling a group to a maximum value of 4 can be more beneficial for more uniformly-distributed data, because the resulting distribution of quantization error becomes more uniform.
Adaptive Block-Scaled Data Types offer each group the choice between standard FP4 and an even more uniform distribution of error in the form of scaled INT4 values, as shown in \cref{fig:if4}.
Like NVFP4, we scale groups of 16 values by an FP8 E4M3 scale factor with 4 exponent bits and 3 significand bits.
The choice of whether a group's values are saved as floats or scaled integers is denoted with the scale factor's sign bit, which is currently unused in NVFP4 since each value has its own sign bit.

We call this data type \textit{IF4} (Int/Float 4, pronounced ``eye-eff-four''), reflecting its ability to switch between these two data types.
However, this method is not limited to 4 bits: in the sections that follow, we additionally propose and analyze \textit{IF3}, which allows values to be stored as E2M0 floating point values or scaled integers, and \textit{IF6}, which leverages the same concept but with 6 bits per value.
We find that not only do Adaptive Block-Scaled Data Types improve quantized training and inference performance, but they can be implemented efficiently in future hardware accelerators with no memory overhead and minimal computational overhead compared to existing block-scaled data types.
\section{Related Work}

Many works have analyzed the performance degradation of large language models (LLMs) that are quantized with 4-bit formats.
When quantizing existing models with post-training quantization (PTQ), much existing literature focuses on W4A16, in which weights are quantized to INT4~\cite{lin_awq_2024,liang_paroquant_2026,frantar_gptq_2023,dettmers_case_2023,kim_squeezellm_2024,tseng_quip_2024} and activations are kept in high precision, or W4A8, in which activations are quantized to INT8 or FP8~\cite{dai_vs-quant_2021,lin_qserve_2025}.
In this work, we focus on the much more challenging W4A4 paradigm, in which weights and activations are quantized to 4 bits.
We additionally study quantized training with W4A4G4, in which weights, activations, and gradients are quantized to 4 bits during training, which has received increased attention due to its support in newer hardware accelerators.

\begin{table}
    \centering
    \begin{threeparttable}
        \caption{Summary of Block-Scaled 4-Bit Quantization Schemes}
        \label{tab:four-bit-datatype-comparison}
        \begin{tabular}{lrrrrrr}
            \toprule
            \multirow{2}{1.0cm}{Format} & \multirow{2}{0.9cm}{\raggedleft Group Size} & \multirow{2}{1.0cm}{\raggedleft Scale Format} & \multirow{2}{1.5cm}{\raggedleft Maximum                                                           \\Value\tnote{\dag}\hspace{0.11cm}} & \multirow{2}{1.8cm}{\raggedleft Minimum\\Value\tnote{\dag}\hspace{0.11cm}} & \multirow{2}{2.3cm}{\raggedleft Relative\\Dynamic Range} & \multirow{2}{1.1cm}{\raggedleft \mbox{MSE\tnote{\ddag}\hspace{0.1cm}}\\$\times$~10$^{-3}$} \\
            \\
            \midrule
            MXFP4                       & 32                                          & UE8M0                                         & 6 $\times$ 2$^{127}$                    & 0.5 $\times$ 2$^{-127}$ & $\infty$       & 13.2         \\
            \midrule
            NVFP4                       & 16                                          & E4M3                                          & 6 $\times$ 448                          & 0.5 $\times$ 2$^{-9}$      & 100\%          & 9.0          \\
            NVFP4 (4/6)                 & 16                                          & E4M3                                          & 6 $\times$ 256                          & 0.5 $\times$ 2$^{-9}$      & 57.1\%         & 7.5          \\
            NVINT4                      & 16                                          & E4M3                                          & 7 $\times$ 448                          & 1 $\times$ 2$^{-9}$        & 58.3\%         & 7.4          \\
            IF4                         & 16                                          & UE4M3                                         & 6 $\times$ 448                          & 0.5 $\times$ 2$^{-9}$      & \textbf{100\%} & \textbf{6.2} \\
            \bottomrule
        \end{tabular}
        \begin{tablenotes}
            \footnotesize
            \item[\dag] Maximum and minimum values are absolute and before application of the FP32 tensor scale factor~\cite{nvidia_pretraining_2026}, which makes the tensor's representable values larger or smaller without affecting its dynamic range.
            \item[\ddag] Mean squared error over $\mathcal{N}(0,1)$.
        \end{tablenotes}
    \end{threeparttable}
\end{table}

\textbf{4-bit numerical formats.}
NVFP4 and MXFP4 have enjoyed increased adoption due to their support in NVIDIA Blackwell and AMD MI355X GPUs.
They have proven to be strong formats for 4-bit quantization, as block-scaled floating point formats are well-suited to represent distributions with outliers~\cite{egiazarian_bridging_2026}.
However, recent works have found that when the distribution of values to quantize is more uniform, such as after undergoing a Hadamard transformation, NVFP4 can underperform relative to other proposed data types such as NVINT4 (an FP8 scale factor for every 16 INT4 values)~\cite{chen_int_2025,egiazarian_bridging_2026}.
Other works have leveraged similar insights to design 4-bit data types with non-standard distributions of values~\cite{guo_ant_2022,dettmers_qlora_2023,dotzel_learning_2024}.

\textbf{Post-training quantization (PTQ) with W4A4.}
W4A4 has received increased attention motivated by its hardware support and the growing memory costs of running frontier language models.
Several techniques for W4A4 quantization with INT4 have been developed using learned rotation matrices~\cite{liu_spinquant_2025,sun_flatquant_2025} or Hadamard transformations~\cite{ashkboos_quarot_2024}.
These techniques generally aim to flatten the distribution of input values, which prior works have shown can perform worse when used with block-scaled FP4 formats such as MXFP4 and NVFP4~\cite{egiazarian_bridging_2026,chen_wush_2026,chen_int_2025}.
Newer techniques such as MR-GPTQ and WUSH rely on block-wise transforms to recover as much performance as possible with these formats~\cite{egiazarian_bridging_2026,chen_wush_2026}.

\textbf{Quantized training with W4A4G4.}
W4A4G4 is a difficult paradigm in which weights, activations, and gradients must all be quantized using 4-bit formats during training.
Existing works tend to keep many operations in high precision, such as the model's forward pass~\cite{tseng_training_2025}, several of the model's hidden layers~\cite{nvidia_pretraining_2026,cook_four_2026}.
Other works introduce operations for suppressing weight oscillation~\cite{chen_tetrajet-v2_2025}, reducing bias in the backward pass~\cite{panferov_quartet_2026}, increase model size to offset the lack of precision~\cite{castro_quartet_2026}, or switch to high-precision at the end of training~\cite{nvidia_nvidia_2025,chmiel_fp4_2025}.
These operations add overhead and may still be unable to match the performance of models trained in high precision, limiting the widespread adoption of W4A4G4.
More capable numerical formats may be needed to make this paradigm viable.
\section{Adaptive Block-Scaled Data Types}
\label{sec:if4}

This section introduces \textit{Adaptive Block-Scaled Data Types} for low-precision training and inference, with a focus on our proposed format for 4-bit quantization, IF4 (Int/Float 4, pronounced ``eye-eff-four'').
However, the same insights can be applied to create formats for other bit-widths: in \cref{sec:discussion}, we also introduce and analyze IF3 and IF6 for 3-bit and 6-bit quantization respectively.

IF4 is inspired by the design of NVFP4~\cite{nvidia_pretraining_2026} and the analysis of Four Over Six (4/6)~\cite{cook_four_2026}.
Like NVFP4, IF4 is a block-scaled format with a tensor-wide FP32 scale factor $\alpha$ and an FP8 E4M3 scale factor $\Delta_i$ for every 16 values.
These can be computed as follows, where $M^\text{FP4}$ and $M^\text{FP8}$ are the largest values that can be represented by FP4 E2M1 and FP8 E4M3, 6 and 448 respectively:

\newcommand{\hp}{\mathbf{X}}
\begin{align}
    \alpha^{\text{FP32}}  & = \frac{\text{max}(|\hp|)}{M^\text{FP4} \times M^\text{FP8}} \label{eqn:fp4-quantization-tensor-scale}   \\
    \Delta^{\text{FP8}}_i & = \frac{\text{max}(|\hp_{16i...16(i+1)}|)}{\alpha M^\text{FP4}} \label{eqn:fp4-quantization-block-scale}
\end{align}

Calculating $\Delta_i$ with $M^\text{FP4}=6$ results in each group's largest value getting quantized to 6, and the remaining values getting scaled to the range of FP4, from 0.5 to 6.
4/6 demonstrated that by doing this, the distribution of FP4 quantization error results in large amounts of quantization error for the near-maximal values in each group of 16 values (\cref{fig:if4}, left)~\cite{cook_four_2026}.
In 4/6, this is partly resolved by quantizing each block twice, once with $M^\text{FP4}=6$ and once with $M^\text{FP4}=4$, and choosing the option with less mean squared quantization error.

\begin{figure}
    \centering
    \begin{subfigure}[b]{0.6\textwidth}
        \centering
        \begin{tabular}{p{0.2cm}p{3.8cm}r}
            \toprule
                             & Pseudocode                                                                                                                      & $\mathbf{X}$ = [6, 18, 36, 42]                           \\
            \midrule
            {\raggedright 1} & $\Delta$ = max($|\mathbf{X}|$) $\div$ 6                                                                                         & 7.0                                                      \\
            2                & $\Delta$ = fp8\_e4m3($\Delta$)                                                                                                  & 7.0                                                      \\
            3                & $\mathbf{\bar{X}}_i$ = $\mathbf{X}_i$ $\div$ $\Delta$                                                                           & [0.86, 2.57, 5.14, 6.0]                                  \\
            4                & $\mathbf{\bar{X}}_i^{(\text{FP})}$ = \Circled{fp4\_e2m1($\mathbf{\bar{X}}_i$)}                                                  & \Circled{[1.0, 3.0, 6.0, 6.0]}                           \\
            5                & $\mathbf{D}_i^{(\text{FP})}$ = $\mathbf{\bar{X}}_i^{(\text{FP})}$ $\times$ $\Delta$                                             & [7.0, 21.0, 42.0, 42.0]                                  \\
            6                & \mbox{$\text{E}^{(\text{FP})}$ = $\frac{1}{n}\sum_i^n(\mathbf{D}_i^{(\text{FP})}-\mathbf{X}_i)^2$}                              & 46.0                                                     \\
            \midrule
            7                & $\mathbf{\bar{X}}_i^{(\text{INT})}$ = \Circled[outer color=teal,fill color=teal]{int4($\mathbf{\bar{X}}_i \times \frac{7}{6}$)} & \Circled[outer color=teal,fill color=teal]{[1, 3, 6, 7]} \\
            8                & $\mathbf{D}_i^{(\text{INT})}$ = $\mathbf{\bar{X}}_i^{(\text{INT})} \times \Delta \times \frac{6}{7}$                            & [6, 18, 36, 42]                                          \\
            9                & \mbox{$\text{E}^{(\text{INT})}$ = $\frac{1}{n}\sum_i^n(\mathbf{D}_i^{(\text{INT})}-\mathbf{X}_i)^2$}                            & 0.0                                                      \\
            \midrule
            10               & \mbox{$\mathbf{\bar{X}}_i =
                    \begin{cases}
                        \mathbf{\bar{X}}_i^{(\text{INT})}, & \text{E}^{(\text{INT})} < \text{E}^{(\text{FP})} \\
                        \mathbf{\bar{X}}_i^{(\text{FP})},  & \text{otherwise}
                    \end{cases}$}
                             & \Circled[outer color=teal,fill color=teal]{[1, 3, 6, 7]}                                                                                                                                   \\
            \bottomrule
        \end{tabular}
        \vspace{0.2cm}
        \caption{Compared to NVFP4, these sample values can be quantized with less error using scaled integers. The standard NVFP4 quantization algorithm ends on line 4, returning $\Delta$ and $\mathbf{\bar{X}}^{(\text{FP})}$. IF4 evaluates both options before returning the representation with less error for each group of 16 values.}
        \label{tab:sample-if4-quantization}
    \end{subfigure}
    \hfill
    \begin{subfigure}[b]{0.38\textwidth}
        \centering
        \includegraphics[width=\linewidth]{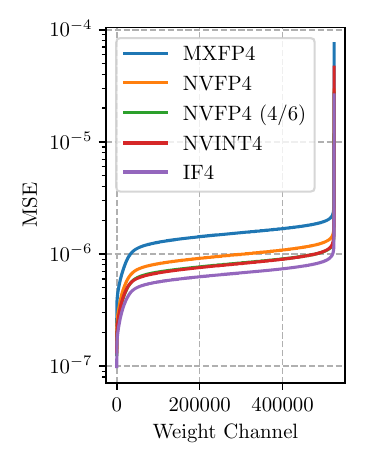}
        \caption{Mean squared quantization error across weight channels for an MoE down projection weight tensor in Qwen3.5-35B-A3B in order of increasing error. Note that NVFP4 (4/6) and NVINT4 have extremely similar error distributions in this figure.}
        \label{fig:quantization-error-distributions}
    \end{subfigure}
    \caption{IF4 quantization is done by quantizing each group to FP4 and INT4 and selecting the option with less mean squared error, resulting in less quantization error across many weight channels in Qwen3.5-35B-A3B with no storage overhead compared to NVFP4.}
\end{figure}

In effect, this allows each group of values the choice between the standard FP4 error distribution, which favors groups with one outlier and many small values, and a slightly more uniform distribution of error, which favors more uniform input distributions.
However, while 4/6 reduces quantization error and empirically helps during quantized training and inference, it comes with two problems:

\begin{enumerate}
    \item Using $M^\text{FP4}=4$ prevents the group from using -6 and 6, giving up two of only 15 useful FP4 values.\footnote{While there are 16 possible FP4 values, one of them is negative zero, which is redundant.}
    \item In order to allow the group with the tensor's largest values to select a maximum value of 4, 4/6 reduces the FP32 tensor scale from $\frac{\text{max}(|\hp|)}{6 \times 448}$ to $\frac{\text{max}(|\hp|)}{6 \times 256}$, reducing the quantized tensor's dynamic range by 42.9\%.
\end{enumerate}

To mitigate the first issue, IF4 replaces the option of scaling to 4 with the option to represent the group's values as INT4, yielding two options with 15 useful values.\footnote{We give up the INT4 value of -8 in order to keep our positive and negative error distributions symmetric.}
To mitigate the second issue, IF4 scales the values in groups that select INT4 by a factor of $\sfrac{7}{6}$ before quantization, and de-scales these values by a factor of $\sfrac{6}{7}$ after de-quantization during matrix multiplication.
Since the maximum representable value in FP4 is 6 and the maximum representable value in INT4 is 7, this results in all groups being quantized to the same range, from 0 to 6 (\cref{fig:if4}).
This is the \textit{6/7 Method of 4-Bit Quantization}.

This procedure is outlined in detail in \cref{tab:sample-if4-quantization}.
When groups are quantized to scaled INT4, the sign bit of the group's scale factor is set to 1 so that its values can be de-quantized and scaled properly later on.
We find that these features result in a data type that has no memory overhead compared to NVFP4, while enjoying significantly reduced quantization error compared to existing block-scaled 4-bit formats.
\cref{tab:four-bit-datatype-comparison} shows that normally distributed values suffer from less mean squared quantization error when quantized with IF4, and \cref{fig:quantization-error-distributions} shows that most weight channels from a weight tensor in Qwen3.5-35B-A3B are represented with less quantization error when quantized with IF4.
\section{Evaluation}

\subsection{Pre-Training with IF4}

\begin{figure}
    \centering
    \includegraphics[width=\textwidth]{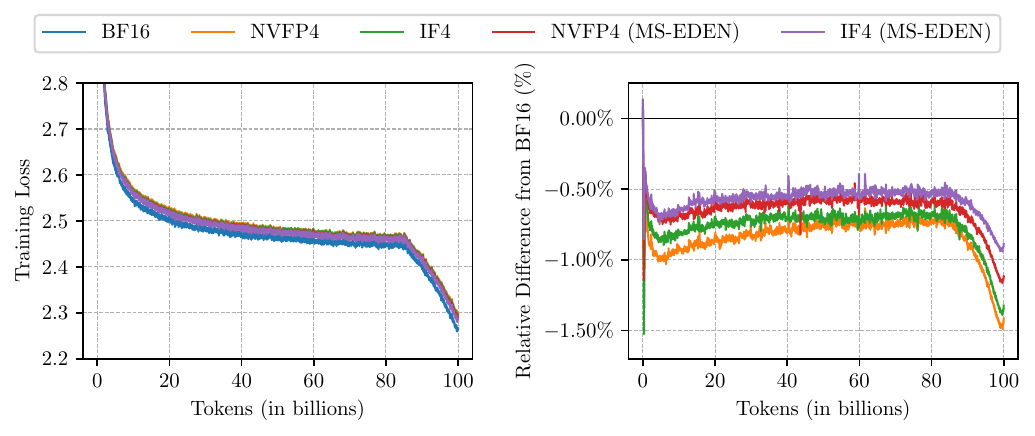}
    \caption{IF4 outperforms NVFP4 during W4A4G4 quantized training. When MS-EDEN~\cite{panferov_quartet_2026} is used in the backward pass, IF4 results in an even larger improvement.}
    \label{fig:if4_mseden}
\end{figure}

We first evaluate the performance of IF4 by pre-training several 340-million-parameter dense Transformer~\cite{vaswani_attention_2023} models with query-key normalization~\cite{henry_query-key_2020} using several different quantization schemes, applying a recipe similar to \cite{nvidia_pretraining_2026} in each experiment.
Specifically, we quantize weights, activations, and gradients in all linear layers except those in the final four hidden layers, we apply the random Hadamard transform to mitigate outliers when computing weight gradients, and we use stochastic rounding when quantizing activation gradients (see \cref{sec:stochastic_rounding} for more details).
All quantization formats studied, including NVFP4, were emulated by de-quantizing to FP32 before converting to BF16 to perform each matrix multiplication.
Following from \cite{panferov_quartet_2026}, we also apply an additional factor of \sfrac{16}{17} during the gradient scale factor calculation to reduce quantization bias that can be introduced by E4M3 scale factors during stochastic rounding.
Our hyper-parameters and training data are described in \cref{sec:training_appendix}.

\begin{figure}
    \centering
    \begin{subfigure}[b]{0.5\textwidth}
        \includegraphics[width=\linewidth]{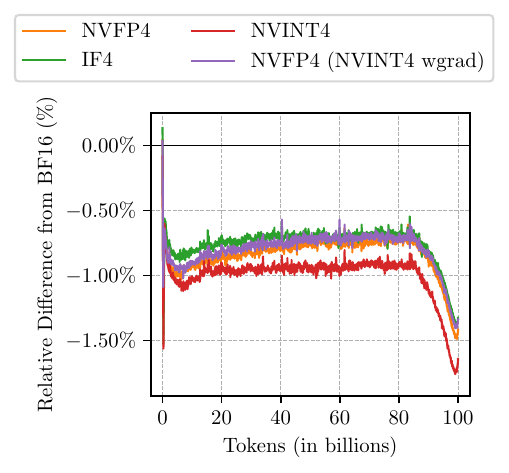}
        \caption{While NVFP4 is strictly better than NVINT4 when used to quantize all tensors, we find that part of the improvement of IF4 can be explained by the ability to represent Hadamard-transformed inputs with less error when computing the weight gradient.}
        \label{fig:wgrad_int4_pretraining}
    \end{subfigure}
    \hfill
    \begin{subfigure}[b]{0.48\textwidth}
        {
            \begin{center}
                \setlength\tabcolsep{5.4pt}
                \begin{tabular}{l|l|ccc|c}
                    \toprule
                                                                                        &        & \multicolumn{3}{c|}{\textbf{Qwen3.5}} & \multirow{2}{*}{Avg.}                                       \\
                                                                                        &        & {\footnotesize 0.8B}                  & {\footnotesize 9B}    & {\footnotesize 27B} &               \\
                    \midrule
                    \multirow{5}{*}{\rotatebox{90}{\parbox{1.82cm}{\vspace{0.2cm}RTN}}} & BF16   & 51.1                                  & 70.5                  & 75.9                & 65.7          \\
                    \midrule
                                                                                        & MXFP4  & 41.4                                  & 66.3                  & 74.2                & 60.6          \\
                                                                                        & NVINT4 & 45.3                                  & 68.7                  & 74.6                & 62.8          \\
                                                                                        & NVFP4  & 45.2                                  & 69.3                  & 75.2                & 63.3          \\
                                                                                        & 4/6    & \textbf{46.7}                         & 69.2                  & 74.9                & 63.6          \\
                                                                                        & IF4    & 46.2                                  & \textbf{70.0}         & \textbf{75.3}       & \textbf{63.8} \\
                    \midrule
                    \multirow{5}{*}{\rotatebox{90}{\parbox{1.82cm}{\centering RTN w/                                                                                                                   \\Hadamard}}} & MXFP4  & 43.4                                  & 67.2                  & 74.7                & 61.6          \\
                                                                                        & NVINT4 & 46.8                                  & \textbf{70.3}         & 74.5                & 63.8          \\
                                                                                        & NVFP4  & 44.4                                  & 67.8                  & 74.3                & 62.3          \\
                                                                                        & 4/6    & 46.3                                  & 68.9                  & 75.1                & 63.5          \\
                                                                                        & IF4    & \textbf{47.6}                         & 69.8                  & \textbf{75.6}       & \textbf{64.3} \\
                    \bottomrule
                \end{tabular}
            \end{center}
        }
        \vspace{0.25cm}
        \caption{When used during W4A4 PTQ, NVINT4 tends to outperform NVFP4 when weights and activations undergo a Hadamard transformation before being quantized. IF4 is able to adapt to raw or transformed inputs, delivering an advantage in both cases.}
        \label{tab:hadamard_ptq}
    \end{subfigure}
    \caption{In W4A4G4 training and W4A4 PTQ, it is common to reduce the dynamic range of the inputs being quantized, such as through a Hadamard transformation. We find that NVINT4 deals with these transformed inputs better, explaining part of IF4's success.}
    \label{fig:hadamard_results}
\end{figure}

We find that IF4 outperforms NVFP4 during pre-training, resulting in training loss closer to a high-precision baseline trained using BF16.
Much of the benefit provided by IF4 can be explained by an increased preference for selecting INT4 in inputs that have undergone a Hadamard transformation.
For example, in \cref{fig:wgrad_int4_pretraining}, we show that IF4 slightly outperforms NVFP4 in our standard pre-training setup.
In another experiment where the NVFP4 recipe is changed such that the weight gradient computation's inputs are quantized using NVINT4, the resulting performance is only marginally worse than IF4.
This finding is strongly supported by \cref{fig:if4_mseden}, which shows that the performance gap between IF4 and NVFP4 is much larger when IF4 is used with MS-EDEN~\cite{panferov_quartet_2026}, which transforms all inputs in the backward pass with a Hadamard transformation.
It is also supported by post-training quantization experiments done on dense models (\cref{tab:hadamard_ptq}), which shows better average performance for NVINT4 if weights and activations are quantized following a Hadamard transformation.

\subsection{Post-Training Quantization with IF4}

{
\setlength\tabcolsep{5pt}
\begin{table}
    \centering
    \begin{tabular}{l|cc|cc|cc|cc}
        \toprule
               & \multicolumn{4}{c|}{\textbf{Nemotron 3}} & \multicolumn{4}{c}{\textbf{Qwen3.5}}                                                                                                                                                                    \\
               & \multicolumn{2}{c|}{\footnotesize 4B}    & \multicolumn{2}{c|}{\footnotesize 30B-A3B} & \multicolumn{2}{c|}{\footnotesize 35B-A3B} & \multicolumn{2}{c}{\footnotesize 122B-A10B}                                                                   \\
               & WikiText-2                                 & C4                                         & WikiText-2                                   & C4                                          & WikiText-2      & C4             & WikiText-2      & C4             \\
        \midrule
        BF16   & 16.92                                    & 43.82                                      & 7.33                                       & 21.75                                       & 7.70          & 22.17          & 5.72          & 21.53          \\
        \midrule
        MXFP4  & 19.77                                    & 51.71                                      & 10.51                                      & 33.17                                       & 8.83          & 25.00          & 6.78          & 23.64          \\
        NVINT4 & 18.35                                    & 47.82                                      & 8.35                                       & 26.32                                       & 8.31          & 23.63          & 6.28          & 22.83          \\
        NVFP4  & 18.03                                    & 47.05                                      & 9.02                                       & 26.86                                       & 8.14          & 23.18          & 6.20          & 22.52          \\
        4/6    & 17.86                                    & 46.68                                      & 9.12                                       & 27.27                                       & 8.11          & \textbf{23.11} & 6.16          & 22.47          \\
        IF4    & \textbf{17.78}                           & \textbf{46.15}                             & \textbf{8.26}                              & \textbf{26.22}                              & \textbf{8.07} & 23.12          & \textbf{6.10} & \textbf{22.42} \\
        \bottomrule
    \end{tabular}
    \vspace{0.2cm}
    \caption{Perplexity results using 4-bit block-scaled quantization schemes with round-to-nearest W4A4 PTQ.}
    \label{tab:perplexity}
\end{table}
}

We also evaluate several models using IF4 during post-training quantization (PTQ).
In all PTQ experiments in this paper, we employ a fairly aggressive paradigm where weights and activations are quantized using round-to-nearest quantization in all linear and Mixture-of-Experts~\cite{shazeer_outrageously_2017} layers except the LM head.

In \cref{tab:perplexity}, we show how model performance is affected by different quantization schemes as measured by WikiText-2 and C4 perplexity for Nemotron 3 Nano~\cite{nvidia_nvidia_2025} and Qwen 3.5~\cite{qwen_team_qwen35_2026} across different model sizes, and find that IF4 outperforms other quantization schemes in nearly all cases.
We use base models to evaluate perplexity when possible: these are provided for Nemotron 3 Nano 30B-A3B and Qwen3.5-35B-A3B.

We also report average model performance on downstream tasks in \cref{tab:tasks}.
To measure performance reliably, we report each combination of model and quantization scheme as an average across five tasks: ARC-Easy, ARC-Challenge~\cite{clark_think_2018}, HellaSwag~\cite{zellers_hellaswag_2019}, LAMBADA~\cite{paperno_lambada_2016}, and PIQA~\cite{bisk_piqa_2019}.
Each task is repeated three times to reduce differences due to nondeterminism.
All experiments are run using vLLM 0.18.0 and lm-eval 0.4.11.
We report \texttt{acc\_norm} for ARC-Easy, ARC-Challenge, HellaSwag and PIQA, and \texttt{acc} from \texttt{lambada\_standard}.
Individual task accuracy scores can be found in \cref{sec:app-ptq}.


{
\begin{table}
    \centering
    \begin{tabular}{l|cc|ccccc|c}
        \toprule
               & \multicolumn{2}{c|}{\textbf{Nemotron 3}} & \multicolumn{5}{c|}{\textbf{Qwen3.5}} & \multirow{2}{*}{Avg. ($\uparrow$)}                                                                                                                         \\
               & {\footnotesize 4B}                       & {\footnotesize 30B-A3B}               & {\footnotesize 9B}                 & {\footnotesize 27B} & {\footnotesize 35B-A3B} & {\footnotesize 122B-A10B} & {\footnotesize 397B-A17B} &               \\
        \midrule
        BF16   & 65.1                                     & 69.4                                  & 70.5                               & 75.9                & 75.5                    & 76.4                      & 81.5                      & 73.5          \\
        \midrule
        MXFP4  & 61.5                                     & 63.8                                  & 66.3                               & 74.2                & 72.8                    & 75.5                      & 79.5                      & 70.5          \\
        NVINT4 & 63.1                                     & 67.5                                  & 68.7                               & 74.6                & 74.0                    & 76.0                      & 80.8                      & 72.1          \\
        NVFP4  & 64.1                                     & 67.6                                  & 69.3                               & 75.2                & 73.9                    & 76.0                      & 80.9                      & 72.4          \\
        4/6    & 63.9                                     & 67.0                                  & 69.2                               & 74.9                & 74.6                    & 75.7                      & 80.8                      & 72.3          \\
        IF4    & \textbf{64.3}                            & \textbf{68.4}                         & \textbf{70.0}                      & \textbf{75.3}       & \textbf{74.6}           & \textbf{76.4}             & \textbf{80.9}             & \textbf{72.8} \\
        \bottomrule
    \end{tabular}
    \vspace{0.2cm}
    \caption{W4A4 round-to-nearest post-training quantization accuracy averaged across five tasks: ARC-Easy, ARC-Challenge, HellaSwag, LAMBADA, and PIQA. Results for individual tasks can be found in \cref{sec:app-ptq}.}
    \label{tab:tasks}
\end{table}
}
\section{Discussion}
\label{sec:discussion}

\subsection{Alternative Data Types}
\label{sec:alternative_data_types}

With an eye toward the future, we simulate NVFP4 and IF4 alongside several other possible block-scaled numerical formats that follow naturally from this work.
These include formats with a smaller block size (i.e. one scale factor for every eight values), and formats for other bit-widths.
Our findings are displayed in \cref{fig:ppl_vs_bits,tab:ppl_vs_bits_table}.
We replicate prior findings of a pareto frontier between model performance and memory~\cite{kumar_scaling_2024}, which we measure using average WikiText-2 perplexity across several different Qwen3.5 model sizes (9B, 27B, 35B-A3B, and 122B-A10B).
We use perplexity for these evaluations as prior works have noted it is often a more reliable metric for evaluating quantized models due to it providing a continuous value per sample~\cite{frantar_gptq_2023,dettmers_case_2023}.
We also find that perplexity is fairly predictive of downstream task performance in our evaluations (\cref{fig:ppl_vs_accuracy}).

Notably, we find that IF (Int/Float) formats offer the best performance for most low-precision memory constraints.
For example, at 4.5 bits per parameter, models quantized to IF4 enjoy lower perplexity than models quantized to NVFP4 and NVINT4, as shown in \cref{tab:perplexity}.
This finding holds at 3.5 bits for IF3\footnote{Note that the 6/7 scaling factor described in \cref{sec:if4} becomes 4/3 for IF3, 7.5/31 for IF6-E2M3, and 28/31 for IF6-E3M2.} versus NVFP3 and NVINT3, and at 4 bits for IF3-BS8 (IF3 with a block size of 8) versus NVFP3-BS8 and NVINT3-BS8.
While it is unclear whether next-generation hardware will be able to efficiently support block sizes smaller than 16, these findings reflect the strengths of Adaptive Block-Scaled Data Types in their ability to deliver high performance in memory-constrained environments.

\subsection{Hardware Support for IF4}

\begin{figure}
    \centering
    \includegraphics[width=\textwidth]{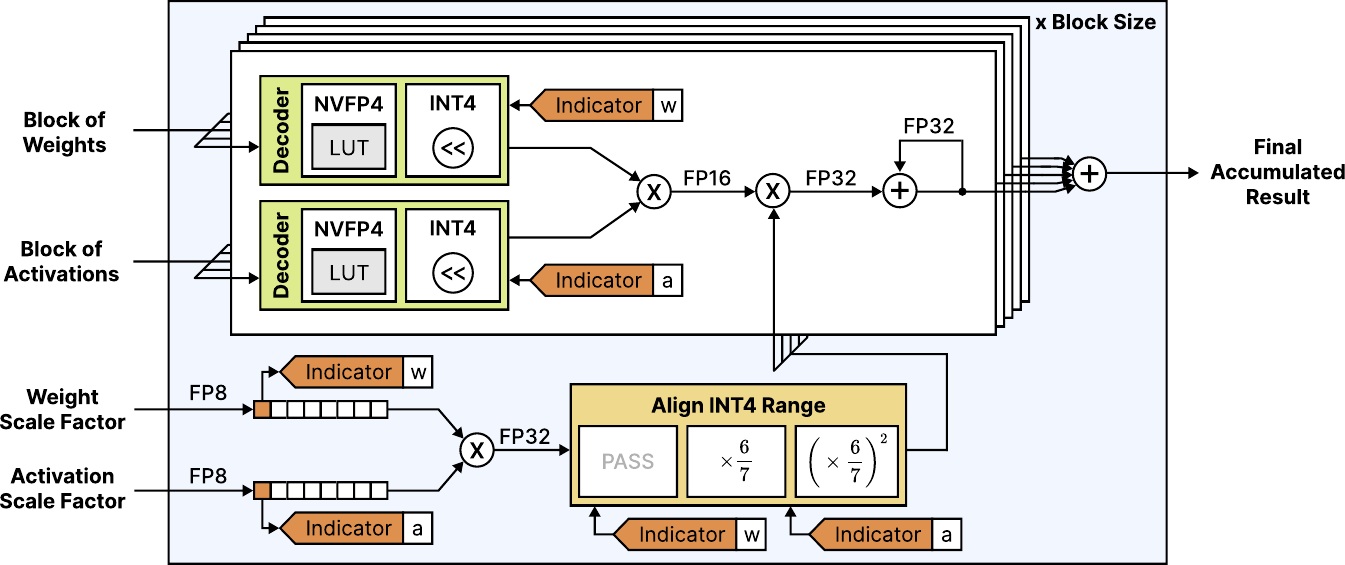}
    \caption{Hardware architecture of an IF4 multiply-accumulate (MAC) block, demonstrating the feasibility of IF4 in next-generation hardware accelerators.}
    \label{fig:if4-mac}
\end{figure}

To evaluate the hardware feasibility of IF4, we implement an IF4 multiply accumulate (MAC) block in SystemVerilog and synthesize it in 28nm CMOS technology.
The MAC takes 16 4-bit weights and 16 4-bit activations, together with the corresponding scale factors for each block.
According to the sign bit of each scale factor, each 4-bit operand is decoded as either NVFP4 or INT4.
For decoding, NVFP4 values are generated through a lookup table (LUT), whereas INT4 values are decoded using simple shifter-based logic.
A LUT-based implementation is well suited for NVFP4 decoding because its value mapping is non-uniform, while the number of distinct magnitude cases is small.

The decoded weight and activation values are first multiplied, and the resulting products are represented in FP16.
In parallel, the weight and activation scale factors are multiplied to produce a unified scale factor.
Handling the scale factors in a separate parallel path avoids repeated scaling operations on individual operands and therefore reduces duplicated computation and hardware cost.
For IF4, additional range-alignment scaling is applied depending on the operand types.
When both operands are decoded as NVFP4, no extra scaling is required.
When one operand is decoded as INT4 and the other as NVFP4, a factor of 6/7 is applied through FP32 multiplication.
When both operands are decoded as INT4, the scaling factor becomes 36/49, also implemented in FP32.
The resulting unified and rescaled scale factor is then broadcast to the 16 element-wise weight-activation products for block scaling.
This stage is implemented with FP16-FP32 scaling and produces FP32 outputs.
Finally, the scaled products are accumulated in FP32 until the accumulator is cleared by the reset signal.

To quantify the hardware overhead of supporting IF4, we implement a plain NVFP4 MAC as a baseline.
Compared to IF4, the NVFP4 MAC uses a simpler datapath in two key aspects: all operands are decoded as NVFP4 through a LUT, and no additional range-alignment scaling is required.
We synthesize the NVFP4 MAC under the same conditions and compare its performance with post-synthesis simulation results.
The detailed architecture of the NVFP4 MAC and full comparison results are provided in \cref{sec:app-hw}.
The latency of IF4 MAC is 4.7\% higher relative to the NVFP4 baseline, primarily due to the additional floating-point multiplication introduced in the range-alignment scaling path.
However, this comparison reflects only the latency of an isolated MAC unit rather than the end-to-end latency of a full system.
In practice, overall throughput is determined by the complete execution pipeline, including operand fetch, data movement, buffering, scheduling, and writeback, so the impact of this additional latency is significantly attenuated at the system level.
Moreover, modern AI workloads on GPUs are often memory-bound, where performance is limited by memory bandwidth rather than compute \cite{recasens2025mind}.
In this regime, small increases in MAC datapath latency have negligible impact on overall throughput, which is instead dominated by HBM bandwidth, cache behavior, and data movement overheads.
\section{Conclusion}

Block-scaled 4-bit numerical formats hold the promise of faster quantized training and inference with minimal performance degradation relative to high-precision models.
However, it remains unclear whether existing formats will be able to deliver these promises due to the lack of precision available when limited to 4 bits per parameter.
In this work, we introduced and evaluated Adaptive Block-Scaled Data Types, which are able to reduce quantization error relative to existing formats while introducing no additional storage overhead.
Our proposed 4-bit format, IF4, achieves lower loss during training and higher accuracy during post-training quantization compared to existing 4-bit quantization schemes.
Furthermore, we showed that it can be implemented efficiently in next-generation hardware accelerators.

\begin{ack}
  We thank Modal and the MIT Office of Research Computing and Data for providing access to computational resources needed to run our experiments.
  This research is partially supported by Amazon, Hyundai Motor Company, the MIT AI Hardware Program, the MIT-IBM Watson AI Lab, the National Science Foundation, the National Science Foundation Graduate Research Fellowship under Grant No. 2141064, and the Advanced Research Projects Agency for Health (ARPA-H) under Award Number D24AC00040-00.
  Any opinion, findings, and conclusions or recommendations expressed in this material are those of the authors and do not necessarily reflect the views of ARPA-H or the National Science Foundation.
\end{ack}

\bibliographystyle{unsrt}
\bibliography{ref,ref_zotero}


\newpage
\crefalias{section}{appendix}
\appendix

\section{Training Setup}
\label{sec:training_appendix}

\begin{table}[h!]
    \centering
    \begin{tabular}{lr}
        \toprule
        \textbf{Hyper-parameter} & \textbf{Value}         \\
        \midrule
        Activation               & Swish                  \\
        AdamW $\beta_1$                & 0.9                    \\
        AdamW $\beta_2$                & 0.95                   \\
        Attention Heads          & 16                     \\
        Context Length           & 8192                   \\
        Gradient Clipping        & 1.0                    \\
        Head Dimension           & 64                     \\
        Hidden Size              & 1024                   \\
        Intermediate Size        & 2816                   \\
        Learning Rate            & 1.2 $\times$ 10$^{-3}$ \\
        Number of Layers         & 24                     \\
        Per-GPU Batch Size & 65536 \\
        Weight Decay             & 0.1                    \\
        \bottomrule
    \end{tabular}
    \vspace{0.2cm}
    \caption{Hyper-parameters used during training.}
    \label{tab:hyperparameters}
\end{table}

Hyper-parameters for all of our training experiments are detailed in \cref{tab:hyperparameters}.
All experiments use the Llama-2 tokenizer~\cite{touvron_llama_2023} with a vocabulary size of 32000 tokens, and we train our models on the FineWeb-Edu dataset~\cite{penedo_fineweb_2024} using AdamW~\cite{loshchilov_decoupled_2019}.
We use Flame~\cite{yang2025flame} to train our models using fully sharded data parallelism and varlen Attention~\cite{vaswani_attention_2023} across eight H100 GPUs.
\section{Quantization Bias with Stochastic Rounding}
\label{sec:stochastic_rounding}

\begin{figure}[h!]
    \centering
    \includegraphics[width=\textwidth]{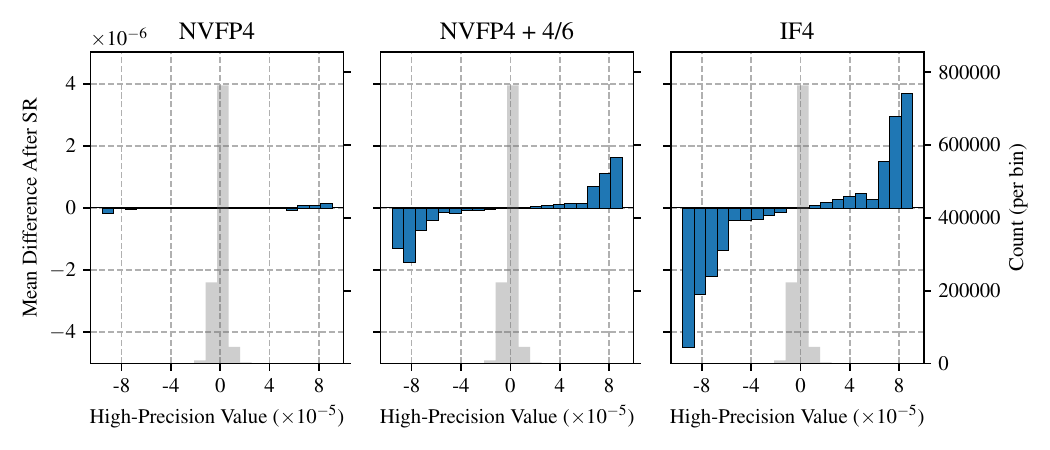}
    \caption{Average change in high-precision values after being quantized to each data type across 1000 trials with stochastic rounding. While large values change the most when quantized with IF4, we find this only affects a very small number of gradients, as most gradients (shown in gray) are significantly smaller than outlier gradient values.}
    \label{fig:stochastic_rounding}
\end{figure}

Unbiased gradient estimation with stochastic rounding has proven to be an essential part of recent successful FP4 training recipes.
While stochastic rounding allows standard NVFP4 to quantize values without bias, 4/6 has been shown to exhibit quantization bias when combined with stochastic rounding due to its preference for selecting values that are closer to representable FP4 values~\cite{cook_four_2026,panferov_quartet_2026}.
Due to its reliance on a similar selection rule, IF4 also features quantization bias when combined with stochastic rounding, which we visualize in \cref{fig:stochastic_rounding} by taking values from a gradient tensor in each of our training runs and plotting the difference in their average values after they are quantized with stochastic rounding 1000 times.
Overall, we find that both 4/6 and IF6 bias larger values toward becoming slightly larger, but that this effect is minimal overall.
While this bias could be mitigated by selecting a data type with round-to-nearest before performing stochastic rounding, this would slow down performance and we have not found it to improve training loss.

We believe the reason it may not improve training loss is likely due to how few values this bias affects.
In \cref{fig:stochastic_rounding}, the distribution of gradient values is shown in gray.
For both 4/6 and IF4, the vast majority of gradient values lie in areas that are mostly unaffected by quantization bias.
Even for values that are affected by this bias, they only change by about 5\% of the magnitude of the largest gradient value in the tensor.
Larger-scale experiments will likely be needed to fully analyze these effects in future work.
\section{Hardware Performance Comparison of NVFP4 and IF4 MACs}
\label{sec:app-hw}

\begin{figure}[h]
    \centering
    \includegraphics[width=\textwidth]{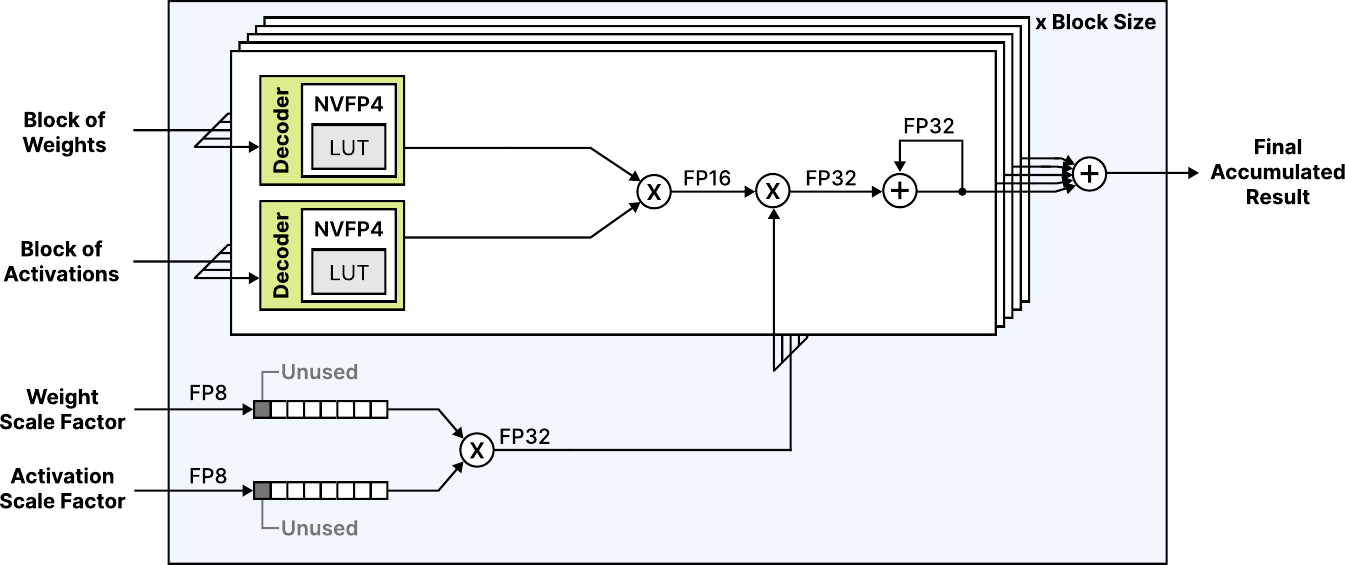}
    \caption{Hardware architecture of NVFP4 multiply-accumulate (MAC) block, providing the baseline for IF4 MAC}
    \label{fig:nvfp4-mac}
\end{figure}

For comparison, we also implemented a plain NVFP4 MAC baseline. Its hardware architecture follows the same overall MAC structure as the IF4 design introduced in the main text, but with a simplified datapath specialized for NVFP4 only. The MAC assumes the same block scaling organization with block size 16 and takes sixteen 4-bit weights and sixteen 4-bit activations together with their corresponding scale factors. Since all operands are interpreted as NVFP4, the decoder does not require mixed-precision selection logic. Each 4-bit operand is directly mapped to its decoded value through an NVFP4 lookup table. After decoding, the weight and activation values are multiplied to generate FP16 partial products. In parallel, the weight and activation scale factors are multiplied to form a unified scale factor for the block. Unlike the IF4 MAC, no additional range-alignment scaling is required. The unified scale factor is broadcast to the 16 element-wise products for block scaling, and the scaled outputs are represented in FP32. Finally, the scaled products are accumulated in FP32 until the accumulator is reset.

{
\setlength\tabcolsep{5.8pt}
\begin{table}[h]
    \centering
    \begin{tabular}{l|cc}
        \toprule
        & \textbf{NVFP4 MAC} & \textbf{IF4 MAC} \\
        \midrule
        Technology (nm) & 28 & 28 \\
        Clock (MHz) & 524 & 500 \\
        Latency (ns) & 3.82 & 4.00 \\
        Throughput (GFLOPS) & 16.77 & 16.00 \\
        Power (mW) & 11.27 & 14.40 \\
        Area (mm$^2$) & 0.024 & 0.040 \\
        Energy Efficiency (GFLOPS/W) & 1488.0 & 1111.1 \\
        Area Efficiency (GFLOPS/mm$^2$) & 698.8 & 400.0 \\
        \bottomrule
    \end{tabular}
    \vspace{0.2cm}
    \caption{Post-synthesis performance comparison between NVFP4 MAC and IF4 MAC}
    \label{tab:mac_comparison}
\end{table}
}
Table \ref{tab:mac_comparison} summarizes the post-synthesis comparison between a plain NVFP4 MAC and the proposed IF4 MAC, implemented under the same conditions. Both designs were synthesized in 28 nm CMOS technology with a target frequency of 500 MHz. The NVFP4 MAC achieved a positive slack of 92 ps, corresponding to a maximum frequency of 524 MHz, while the IF4 MAC met timing at 500 MHz with zero slack. Latency is defined as the time required to complete one MAC operation, and throughput, power, and area are reported per MAC. The NVFP4 MAC achieves a latency of 3.82 ns at 524 MHz, corresponding to 16.77 GFLOPS, while the IF4 MAC operates at 500 MHz with a latency of 4.00 ns and 16.00 GFLOPS. This corresponds to a 4.7\% increase in latency and a 4.6\% reduction in throughput for IF4. In terms of cost, IF4 shows a 27.8\% increase in power and a 66.6\% increase in area, resulting in reductions of 25.3\% in energy efficiency and 42.8\% in area efficiency. At first glance, these overheads may appear significant. However, this comparison isolates a minimal MAC datapath rather than reflecting a full accelerator or even a complete processing element (PE). As a result, the additional logic introduced by IF4 is disproportionately amplified in this setup.

The observed area and power increases for IF4 can be traced to two concrete additions in the MAC datapath. First, decoded 4-bit operands are stored in slightly wider fixed-point formats (Q3.1 for NVFP4 versus Q4.1 for IF4), which marginally widens the datapath and associated storage. Second, IF4 introduces an extra floating-point multiplier to perform range‑alignment scaling, a function that is entirely absent in the plain NVFP4 baseline, which is specialized to a single format and thus implements a highly simplified datapath. In other words, the reported overhead reflects the cost of moving from a minimal single‑format MAC to a mixed‑precision IF4 MAC, not the incremental cost of adding IF4 into an already multi‑format compute block; in realistic GPUs, mixed‑precision tensor/matrix units already include additional reconfigurable logic and datapath resources to support multiple precisions and formats. At the PE level, the MAC datapath typically accounts for only a fraction of the total area, which is dominated by register files, interconnect, control logic, local buffering, and auxiliary functions such as quantization or pooling \cite{chen2016eyeriss}. In such a context, slightly wider internal datapaths and one extra floating‑point multiplier constitute a small perturbation relative to the overall arithmetic and storage footprint of the PE, so the apparent 66.6\% MAC‑level area increase substantially overstates the hardware‑level impact. Moreover, numerous studies of DNN accelerators show that energy and power are largely dominated by memory accesses and data movement rather than MAC operations, with DRAM and on‑chip memory  often accounting for the majority of the energy in both inference and training workloads \cite{chen2016eyeriss, horowitz20141, sze2017efficient}. Consequently, the incremental power attributable to IF4‑specific logic is expected to have limited impact on overall system power and end‑to‑end efficiency once embedded in a realistic PE and accelerator where memory traffic, dataflow, and buffering behavior dominate the energy budget.

\section{Post-Training Quantization}
\label{sec:app-ptq}

\subsection{Perplexity}

\begin{figure}
    \includegraphics[width=\linewidth]{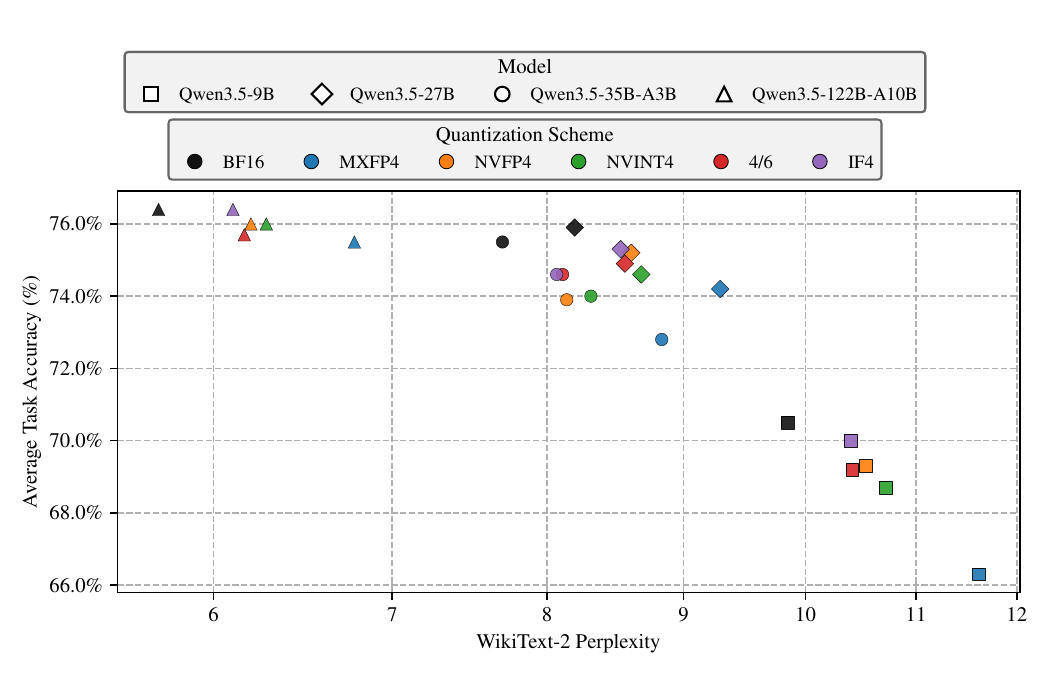}
    \caption{We find WikiText-2 Perplexity to be a fairly strong indicator of average downstream task accuracy across a variety of Qwen3.5 model sizes. This helps us simulate future data types in \cref{sec:alternative_data_types} while using fewer resources.}
    \label{fig:ppl_vs_accuracy}
\end{figure}

\begin{table}
    \centering
    \begin{tabular}{lllrrr}
        \toprule
        \multirow{2}{*}{Name} & Scale Factor & Value       & Block & Bits per  & Avg. WikiText-2 \\
                              & Type         & Type        & Size  & Parameter & Perplexity      \\
        \midrule
        BF16                  & --           & --          & --    & 16.0      & 7.87            \\
        \midrule
        MXFP3                 & UE8M0        & E2M0        & 32    & 3.25      & 70.04           \\
        NVINT3                & E4M3         & INT3        & 16    & 3.5       & 12.78           \\
        NVFP3                 & E4M3         & E2M0        & 16    & 3.5       & 11.12           \\
        IF3                   & UE4M3        & INT3 / E2M0 & 16    & 3.5       & 10.67           \\
        NVINT3-BS8            & E4M3         & INT3        & 8     & 4.0       & 10.27           \\
        NVFP3-BS8             & E4M3         & E2M0        & 8     & 4.0       & 9.74            \\
        IF3-BS8               & UE4M3        & INT3 / E2M0 & 8     & 4.0       & 9.37            \\
        \midrule
        MXFP4                 & UE8M0        & E2M1        & 32    & 4.25      & 9.13            \\
        NVINT4                & E4M3         & INT4        & 16    & 4.5       & 8.50            \\
        NVFP4                 & E4M3         & E2M1        & 16    & 4.5       & 8.37            \\
        NVFP4 + 4/6           & E4M3         & E2M1        & 16    & 4.5       & 8.31            \\
        IF4                   & UE4M3        & INT4 / E2M1 & 16    & 4.5       & 8.27            \\
        NVINT4-BS8            & E4M3         & INT4        & 8     & 5.0       & 8.24            \\
        NVFP4-BS8             & E4M3         & E2M1        & 8     & 5.0       & 8.22            \\
        NVFP4-BS8 + 4/6       & E4M3         & E2M1        & 8     & 5.0       & 8.31            \\
        IF4-BS8               & UE4M3        & INT4 / E2M1 & 8     & 5.0       & 8.11            \\
        \midrule
        MXFP6-E2M3            & UE8M0        & E2M3        & 32    & 6.25      & 7.93            \\
        MXFP6-E3M2            & UE8M0        & E3M2        & 16    & 6.25      & 8.03            \\
        NVINT6                & E4M3         & INT6        & 16    & 6.5       & 7.91            \\
        NVFP6-E2M3            & E4M3         & E2M3        & 16    & 6.5       & 7.91            \\
        NVFP6-E3M2            & E4M3         & E3M2        & 16    & 6.5       & 7.97            \\
        IF6-E2M3              & UE4M3        & INT6 / E2M3 & 16    & 6.5       & 7.90            \\
        IF6-E3M2              & UE4M3        & INT6 / E3M2 & 16    & 6.5       & 7.90            \\
        \bottomrule
    \end{tabular}
    \vspace{0.2cm}
    \caption{Average WikiText-2 perplexity across Qwen3.5-9B, Qwen3.5-27B, Qwen3.5-35B-A3B, and Qwen3.5-122B-A10B when weights and activations in all linear and MoE layers but the LM head are quantized using the specified quantization scheme.}
    \label{tab:ppl_vs_bits_table}
\end{table}

In \cref{tab:ppl_vs_bits_table}, we report the results displayed visually in \cref{fig:ppl_vs_accuracy}.
ExMy denotes a floating point data type with x exponent bits and y mantissa bits.
U denotes types without a sign bit, meaning that the type can only represent positive values.

\subsection{Downstream Tasks}

In \cref{tab:all-ptq-results}, we report all individual task results from \cref{tab:tasks}.
Each result is reported as an average across three runs, and bold denotes the best quantized result for that combination of model and task.
We plot these results against average WikiText-2 perplexity in \cref{fig:ppl_vs_accuracy} and find that WikiText-2 perplexity is a fairly strong indicator of downstream task accuracy.

{
    \begin{table}
        \centering
        \begin{tabular}{l|l|rrrrr|r}
            \toprule
            Model & Scheme & ARC-E       & ARC-C  & HellaSwag      & LAMBADA        & PIQA           & Average ($\uparrow$)           \\
            \midrule
            \multirow{6}{*}{\hspace{0.15cm}\rotatebox{90}{\parbox{2.21cm}{\centering Nemotron 3                                  \\Nano 4B}}} & BF16 & 79.00 & 53.36 & 67.33 & 49.02 & 76.66 & 65.07 \\
                  & MXFP4  & 75.35          & 49.32          & 63.97          & 43.94          & 75.06          & 61.53          \\
                  & NVINT4 & 76.67          & 50.09          & 65.65          & 46.78          & 76.35          & 63.11          \\
                  & NVFP4  & 77.01          & \textbf{53.73} & \textbf{66.36} & 48.13          & 75.41          & 64.13          \\
                  & 4/6    & \textbf{77.82} & 52.62          & 66.32          & 46.99          & 75.97          & 63.94          \\
                  & IF4    & 77.64          & 52.36          & 66.33          & \textbf{48.66} & \textbf{76.39} & \textbf{64.27} \\
            \midrule
            \multirow{6}{*}{\hspace{0.15cm}\rotatebox{90}{\parbox{2.21cm}{\centering Nemotron 3                                  \\Nano 30B-A3B}}} & BF16 & 75.83& 47.53& 77.90 & 63.71& 81.94& 69.38 \\
                  & MXFP4  & 74.75          & 47.58          & 71.51          & 45.78          & 79.56          & 63.84          \\
                  & NVINT4 & 75.00          & 46.99          & 75.41          & 59.49          & 80.54          & 67.49          \\
                  & NVFP4  & 73.48          & 46.67          & \textbf{77.39} & 59.96          & 80.65          & 67.63          \\
                  & 4/6    & 74.47          & 47.44          & 76.84          & 55.49          & \textbf{80.99} & 67.05          \\
                  & IF4    & \textbf{75.39} & \textbf{48.12} & 77.14          & \textbf{60.93} & 80.32          & \textbf{68.38} \\
            \midrule
            \multirow{6}{*}{\hspace{0.15cm}\rotatebox{90}{\parbox{2.21cm}{\centering Qwen3.5                                     \\9B}}} & BF16 & 74.34& 55.77& 78.04& 64.48& 80.11& 70.55\\
                  & MXFP4  & 72.66          & 50.74          & 74.50           & 56.58          & 77.04          & 66.30           \\
                  & NVINT4 & 73.22          & 53.87          & 76.56          & 60.84          & 78.82          & 68.66          \\
                  & NVFP4  & 74.28          & \textbf{54.55} & 76.68          & 62.92          & 78.26          & 69.34          \\
                  & 4/6    & 74.51          & 53.33          & 76.96          & 62.19          & 79.2           & 69.24          \\
                  & IF4    & \textbf{75.52} & 54.41          & \textbf{77.02} & \textbf{63.22} & \textbf{79.67} & \textbf{69.97} \\
            \midrule
            \multirow{6}{*}{\hspace{0.15cm}\rotatebox{90}{\parbox{2.21cm}{\centering Qwen3.5                                     \\27B}}} & BF16 & 79.67& 61.63& 83.39& 72.47& 82.19& 75.87 \\
                  & MXFP4  & 78.79          & 60.21          & 81.40           & 68.94          & 81.41          & 74.15          \\
                  & NVINT4 & 78.76          & 58.93          & 82.78          & 70.73          & 81.57          & 74.56          \\
                  & NVFP4  & 78.94          & \textbf{61.49} & 82.53          & 70.98          & 82.06          & 75.20          \\
                  & 4/6    & 78.98          & 59.93          & 82.63          & \textbf{71.05} & \textbf{82.15} & 74.95          \\
                  & IF4    & \textbf{80.42} & 60.21          & \textbf{82.95} & 70.75          & \textbf{82.15} & \textbf{75.30} \\
            \midrule
            \multirow{6}{*}{\hspace{0.15cm}\rotatebox{90}{\parbox{2.21cm}{\centering Qwen3.5                                     \\35B-A3B}}} & BF16 & 79.14& 62.32& 82.51& 70.23& 83.19& 75.48\\
                  & MXFP4  & 77.41          & 59.81          & 79.98          & 65.46          & 81.21          & 72.78          \\
                  & NVINT4 & 79.38          & 60.04          & 81.34          & 67.32          & 81.85          & 73.98          \\
                  & NVFP4  & 78.93          & 59.64          & 81.31          & 67.59          & 82.06          & 73.91          \\
                  & 4/6    & 79.77          & \textbf{61.60} & \textbf{81.79} & 67.25          & \textbf{82.70} & 74.62          \\
                  & IF4    & \textbf{80.37} & 60.81          & 81.75          & \textbf{68.04} & 82.17          & \textbf{74.63} \\
            \midrule
            \multirow{6}{*}{\hspace{0.15cm}\rotatebox{90}{\parbox{2.21cm}{\centering Qwen3.5                                     \\122B-A10B}}} & BF16 &80.39& 63.05& 85.92& 69.23& 83.28& 76.37 \\
                  & MXFP4  & 78.90          & 63.25          & 84.76          & 68.21          & 82.35          & 75.50           \\
                  & NVINT4 & 79.70          & 63.57          & 84.77          & 69.08          & 82.72          & 75.97          \\
                  & NVFP4  & 79.94          & 62.37          & 85.00          & \textbf{69.86} & 82.73          & 75.98          \\
                  & 4/6    & 79.78          & 62.23          & 84.85          & 69.51          & 82.28          & 75.73          \\
                  & IF4    & \textbf{80.40} & \textbf{63.85} & \textbf{85.02} & 69.73          & \textbf{82.75} & \textbf{76.35} \\
            \midrule
            \multirow{6}{*}{\hspace{0.15cm}\rotatebox{90}{\parbox{2.21cm}{\centering Qwen3.5                                     \\397B-A17B}}} & BF16 &83.92& 68.71& 91.25& 79.21& 84.64& 81.55 \\
                  & MXFP4  & 83.57          & 64.79          & 89.03          & 77.44          & 82.75          & 79.52          \\
                  & NVINT4 & 83.78          & 67.01          & 90.52          & \textbf{78.80} & 83.87          & 80.79          \\
                  & NVFP4  & 83.54          & \textbf{67.53} & 90.67          & 78.47          & 84.28          & 80.90          \\
                  & 4/6    & 83.94          & 67.04          & 90.60          & 78.52          & 83.84          & 80.79          \\
                  & IF4    & \textbf{84.22} & 66.98          & \textbf{90.85} & 78.12          & \textbf{84.53} & \textbf{80.94} \\
            \bottomrule
        \end{tabular}
        \vspace{0.2cm}
        \caption{Full downstream task results for each combination of model and task from \cref{tab:tasks}. Each experiment is repeated three times and reported as an average across those three repetitions. Bold denotes the best quantized result.}
        \label{tab:all-ptq-results}
    \end{table}
}



\end{document}